\documentclass[letterpaper,12pt]{report}	

\usepackage{styles/trformat}  		

\usepackage{pdflscape}
\usepackage{graphicx}
\usepackage{tabulary}
\usepackage{amsmath,amsthm,amsfonts,amscd}
\usepackage{eucal} 	 	
\usepackage{verbatim}      	
\usepackage{styles/citesort}         	%
\usepackage{algorithm}
\usepackage{bibentry}
\usepackage{algorithmic}
\usepackage{xcolor}
\usepackage{markdown}
\usepackage{booktabs}
\usepackage{url}
\usepackage{rotating}
\usepackage[numbers]{natbib}

\usepackage{multirow}
\usepackage{makecell}
\usepackage{tabularx}
\usepackage{bbm}
\usepackage{mathtools}
\usepackage{hyperref}
\hypersetup{colorlinks=false}
\usepackage{svg}
\usepackage{xstring}

\oneandonehalfspacing

\theoremstyle{definition}

\theoremstyle{remark}

\newcommand{\latexe}{{\LaTeX\kern.125em2%
                      \lower.5ex\hbox{$\varepsilon$}}}

\chardef\bslash=`\\	

\makeatletter		
\def\square{\RIfM@\bgroup\else$\bgroup\aftergroup$\fi
  \vcenter{\hrule\hbox{\vrule\@height.6em\kern.6em\vrule}%
                                              \hrule}\egroup}
\makeatother		

\nobibliography*

\author{Madison Threadgill, Andreas Gerstlauer}  	

\address{Electrical and Computer Engineering\\
         The University of Texas at Austin}  

\date{May 1, 2024}

\title{A Survey of Distributed Learning in Cloud, Mobile, and Edge Settings} 

\trnumber{UT-CERC-24-01}

\supervisor{Andreas Gerstlauer}

\committeemembers
	{Radu Marculescu}

\masterreport

\begin{document}






\setcounter{page}{1}

\trabstract
In the era of deep learning (DL), convolutional neural networks (CNNs), and large language models (LLMs), machine learning (ML) models are becoming increasingly complex, demanding significant computational resources for both inference and training stages. To address this challenge, distributed learning has emerged as a crucial approach, employing parallelization across various devices and environments.  This survey explores the landscape of distributed learning, encompassing cloud and edge settings. We delve into the core concepts of data and model parallelism, examining how models are partitioned across different dimensions and layers to optimize resource utilization and performance.
We analyze various partitioning schemes for different layer types, including fully connected, convolutional, and recurrent layers, highlighting the trade-offs between computational efficiency, communication overhead, and memory constraints. This survey provides valuable insights for future research and development in this rapidly evolving field by comparing and contrasting distributed learning approaches across diverse contexts.

\tableofcontents   

\listoftables      
\listoffigures     

\newpage

%
%

\chapter{Introduction}
With the rise of deep learning, convolutional neural networks (CNNs), and large language models (LLMs), machine learning (ML) models are increasing in computational complexity during both the inference and training stages. 

Parallelization methods have been introduced to overcome the computational cost associated with ML models. There are many different granularities as to which ML models can be parallelized. Fine-grained parallelism can occur in shared memory systems where operational or thread-level parallelism is exploited. This form of parallelism is largely well understood with standard approaches being implemented in most systems today. On the other hand, coarse-grained parallelism can be achieved by distributing the ML model across various devices. This form of parallelism introduces challenges as explicit partitioning of data and/or models must be maintained in a distributed memory fashion. Moreover, the partitioning method must also keep in mind the communication overhead between devices to maintain the performance of the system. Additionally, when implementing coarse-grained partitioning, multiple options arise when determining the system's architecture, such as involving the cloud for computational offloading or keeping all data on edge and/or mobile devices.

In the cloud, partitioning of an ML model is typically implemented to mitigate significant computational costs associated with training. The training process can be distributed across multiple CPU or GPU nodes in a cloud or data center cluster. By contrast, when moving computations towards the edge, where resources are limited, inference tasks are commonly used instead of training due to a decreased computational complexity. Moreover, implementing partitioning methods that specifically account for memory and communication impacts is crucial for fast processing of inference tasks.

Edge and mobile devices are resource constrained with limited computational and communication capabilities. Memory constraints make fitting an entire ML model on a single edge device often impossible. Hence, clusters of edge and mobile devices are used, and the ML model is partitioned across them. Communication is much more expensive in edge clusters than in clusters within the cloud. Therefore, when partitioning an ML model between mobile and edge devices, there is a trade-off between computational capabilities and communication. At the same time, when parallelizing a model exclusively between edge and mobile devices, input data to the ML model can be kept on the device where the data was collected. This partitioning of input data can ensure the privacy of collected data as data is not transmitted to different devices.

Data parallelism \cite{joshi_enabling_2023} is partitioning across the input data dimension, where copies of the entire neural network are placed on multiple devices. Each device then processes subsets of the input data. Federated learning (FL) implements data parallelism in the training process while protecting private data by keeping input data on the edge and mobile devices on which the data is collected. Beyond standard data parallelism, the critical concept of FL revolves around a trade-off between communication efficiency and model accuracy. In FL devices can communicate continuously or at reduced intervals. Continuous communication keeps the model up-to-date. By contrast, more infrequent communication reduces the rate of convergence or, for the same amount of training, reduces accuracy due to outdated or incomplete data. To balance these factors, FL optimizes communication frequency to maintain accuracy, minimize overhead, and preserve data privacy \cite{li2020federated}.

In many mobile and edge scenarios, an entire ML model cannot fit on one device, and model parallelism is implemented, where the neural network is partitioned into sub-models, with each part of a model being placed on a separate device. This allows different parts of a model to be processed in parallel during either inference or training but requires communication of intermediate and internal data at the interface between partitions. Model and data parallelism can be combined, creating an ample design space for coarse-grain distributed ML. 

The rest of this survey is structured as follows: Section 2 will focus on typical layer-based ML model architectures along with model and data parallelism and their respective partitioning schemes. Section 3 in turn will explore different layer types in ML models and how those layers are grouped and partitioned. Section 4 will focus on challenges and future directions related to partitioning schemes. Finally, we will conclude in Section 5.

\chapter{Data and Model Partitioning}

\begin{figure}
\centering
\includegraphics[width=.8\textwidth]{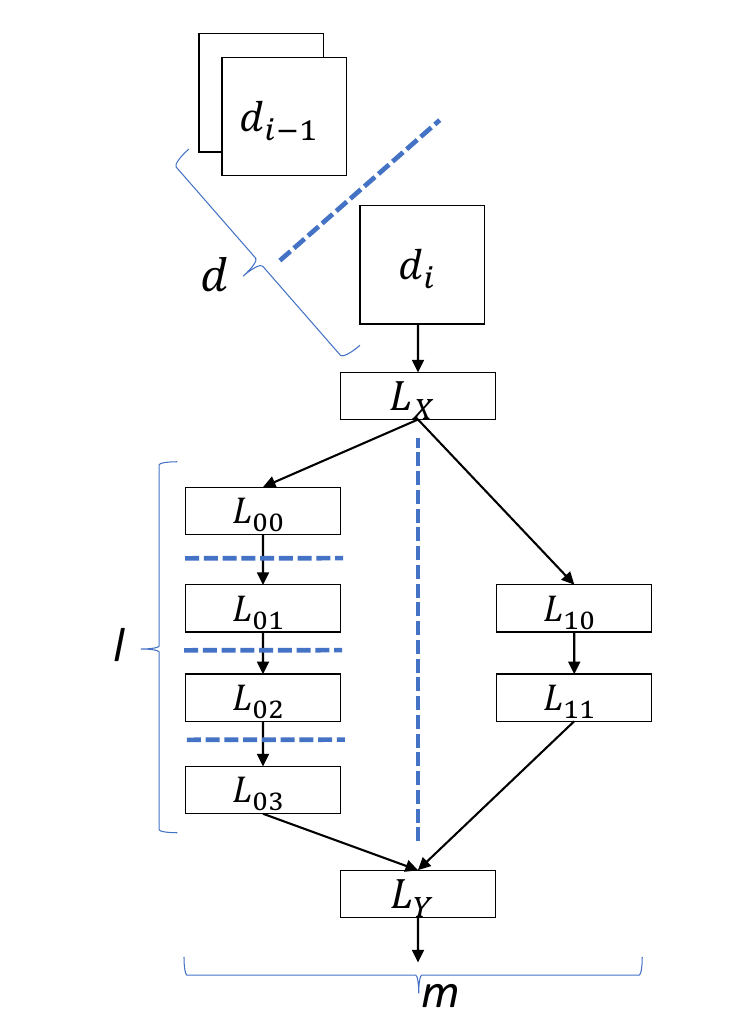}
\caption[Data and model parallelism in a neural network.]{Data and model parallelism in a neural network.}
\label{fig:model_layer}
\end{figure}

This chapter aims to show how a typical ML model architecture can be partitioned to exploit data and model parallelism at the general model architecture level, leaving finer-grained partitioning methods to be discussed in the following chapter. 

Figure \ref{fig:model_layer} illustrates how a simple model can be partitioned within the data (\emph{d}), model path/branch (\emph{m}), and layer (\emph{l}) dimensions. Assuming there are \emph{i} pieces of input data, we can partition across the data (\emph{d}) dimension by sending a specific amount of data samples to one device and the rest of the data samples to a similar model on another device. Neural network models are typically organized as a sequence of layers \emph{$L_X$}...\emph{$L_Y$}, where multiple layers exist in ML networks, with branches between layers, spanning the model partitioning design space. For example, a partitioning configuration can determine that layers \emph{$L_{00}$, $L_{01}$, $L_{02}$, $L_{03}$} can be executed on one device while layers \emph{$L_{10}$} and \emph{$L_{11}$} are executed on another device in parallel. The output of both sets of layers is then sent to \emph{$L_Y$}. Even in this simple model illustrating model parallelism, the choice of partitioning methods is non-trivial and offers a diverse design space. For instance, instead of co-locating layers \emph{$L_{00}$, $L_{01}$, $L_{02}$, $L_{03}$} on a single device, each layer could be allocated to a separate device. Moreover, the size of the design space increases as partitioning schemes differ depending on whether the model is being used for inference or training. With inference, devices may not have to communicate if the data is local. Whereas for training, communication must happen when gradients are updated; moreover, as in FL, the frequency of gradient updates must also be decided.

\begin{table*}[ht]
\caption[Model and data parallelism in ML networks.]{Model and data parallelism in ML networks.} 
\makebox[\linewidth]{
\begin{tabulary}{.8\linewidth}{ c | c | c | c | c | c | c | c} \makecell{} & \makecell{Inf./\\Train.}& \makecell{Cloud/\\Edge} & \makecell{part. \\dim.} & \makecell{comp. \\benefit} & \makecell{comm. \\req.} & \makecell{memory \\benefit} & \makecell{privacy}\\
\hline
\hline
\makecell{Cloud \\ Training \\ \cite{krizhevsky2012imagenet, jia2018beyond, narayananEfficientLargeScaleLanguage2021}} & Train. & \makecell{Cloud} & $d/l/m$ & \makecell{throughput} & \makecell{weights/\\data} & \makecell{weights}  &- \\
\hline
\makecell{Parameter\\Server\\ \cite{liScalingDistributedMachine2014,kimSTRADSDistributedFramework2016}} & Train. & \makecell{Cloud/\\Edge} & $d$/$l$/$m$ &  latency & \makecell{weights} & \makecell{weights/\\data} & - \\
\hline 
\makecell{Cloud-Assisted\\Inference\\ \cite{teerapittayanonDistributedDeepNeural2017,mohammedDistributedInferenceAcceleration2020}} & Inf. & \makecell{Cloud/\\Edge} & $d$/$l$/$m$ & throughput & \makecell{weights/\\data} & \makecell{weights} & x \\
\hline
\makecell{Federated \\ Learning \\ \cite{mcmahan17a, wangWhenEdgeMeets2018, mcmahan2021advances, li_federated_2020, noauthor_coopfl_nodate}} & Train. & \makecell{Cloud/\\Edge} & $d$ & \makecell{throughput} & weights &  - & x\\
\hline
\makecell{Edge\\Inference \\ \cite{hadidi2020toward, zhou2019adaptive, xueEdgeLDLocallyDistributed2020, huai_towards_2019, tang_low-memory_2021, zhou_aaiot_2019, hadidiDistributedPerceptionCollaborative2018}} & Inf. & Edge & $d/l$/$m$ & throughput & \makecell{weights/\\data} & weights & x \\
\hline
\end{tabulary} 
}
\label{data_model_table}
\end{table*}

Table \ref{data_model_table} surveys existing approaches exploring various forms of model and data parallelism. We categorize works based on training or inference contexts, cloud or edge computing environments, partitioning dimensions, computational benefits, communication requirements, memory advantages, and privacy considerations. 

Traditionally, powerful machines such as GPUs are used to train complex models in cloud computing by partitioning the training data batches onto different GPUs in a cluster in a data-parallel fashion, i.e., partitioning in the $d$ dimension \cite{6817512}. This method usually involves copies of the entire model to be placed on different machines, with each worker aggregating its gradients during training until model convergence is achieved \cite{narayananEfficientLargeScaleLanguage2021}. More recently, finer ML model partitioning has been implemented in cloud contexts to train ML models. Partitioning along the $l$ and $m$ dimensions allow parts of the model to be offloaded to devices that can handle the task's computational complexity and increase the system's throughput \cite{krizhevsky2012imagenet}. More recently, works such as \cite{jia2018beyond} have increased the search space to partitioning in the $d$, $l$, and $m$ as well as intra-layer partitioning, introducing algorithms to determine the most efficient partitioning scheme based on each device in the cloud cluster's computational capabilities and the communication latency of the system. 

To further decrease communication overheads and increase parallelism, synchronization requirements are relaxed in the parameter server model during cloud training. For example, the work in \cite{liScalingDistributedMachine2014} introduces asynchronous communication between worker nodes and a server. In this case, the worker nodes collect data and process parts of the ML model. Simultaneously, the server tracks globally shared parameters independently, which helps decrease the system's latency by enabling concurrent execution and avoiding the need for constant synchronization. However, determining when to update the model parameters and distribute these parameters back to worker nodes is a non-trivial problem as parameters may have dependencies and different convergence rates, further increasing the complexity of the design space for training on cloud and edge devices \cite{kimSTRADSDistributedFramework2016}. 

As the popularity of edge computing has increased, many works are leveraging the power of the cloud and edge devices to jointly perform inference tasks. In addition to addressing resource constraints on the edge, the system's throughput increases as the inference process is pipelined between edge devices and the cloud. Since input data is stored on the device it was collected on, with only model features being sent to the cloud for processing, privacy can be protected with cloud-assisted inference \cite{teerapittayanonDistributedDeepNeural2017}. However, communication time remains an issue as input and output data must be communicated between devices and the cloud. To mitigate the communication costs incurred by offloading computation to the cloud works such as \cite{mohammedDistributedInferenceAcceleration2020} further partition the ML network based on each device's computational capabilities to increase the system's throughput.

FL is a machine learning approach where a centralized model, usually located in the cloud, is trained collaboratively across decentralized edge devices, allowing for privacy-preserving and efficient model training without centralized data aggregation. Each edge device holds a local model that is then trained on the input data received on each device. Finally, after a specified period, each edge device sends its updated model weights to the cloud to be aggregated. After this aggregation occurs, the updated weights are sent back to each edge device in the cluster, and another iteration begins \cite{mcmahan17a}. Privacy is maintained as data is kept on the device on which it was collected, and this input data is not sent to the central cloud server. In addition to privacy, FL offers the benefit of increased system throughput due to devices working in parallel. Although this partitioning approach has its advantages, it may overlook the fairness of data distribution across the device cluster, resulting in statistical heterogeneity that could increase the convergence time of the model \cite{li_federated_2020, mcmahan2021advances}. However, communication costs are significant as weight updates must be communicated between the cloud server and each edge device. Works such as \cite{wangWhenEdgeMeets2018} and \cite{noauthor_coopfl_nodate} aim to increase the convergence rate and increase throughput by determining when to perform the global parameter aggregation as well as taking into account the heterogeneity of the system.

Finally, pure edge inference aims to keep all inference tasks on edge devices. This ensures full data privacy. This partitioning scheme typically keeps input data on the device it is collected on while partitioning the model across the edge cluster along the layer-wise or per-branch dimensions based on each device's computational and memory constraints \cite{hadidiDistributedPerceptionCollaborative2018, zhou2019adaptive, hadidi2020toward}. Moreover, approaches in this space also account for the fact that some devices may be idle while other devices have a large workload; therefore, tasks on a single device can exploit the idle computational power of other devices in the network \cite{huai_towards_2019}. Additionally, in this space, some works focus on memory benefits during partitioning \cite{tang_low-memory_2021}, while other works focus on system throughput and data transmission \cite{zhou_aaiot_2019, xueEdgeLDLocallyDistributed2020}.

In summary, data and model parallelism provide practical strategies for scaling DL tasks while optimizing computational resources. A vast amount of literature in the field details various partitioning schemes for data and model parallelism with different advantages and disadvantages. However, achieving finer control and customization in resource allocation within ML networks requires partitioning at the individual layer level. This approach enables tailored optimization to accommodate diverse computational demands and constraints across DL environments and will be discussed in the next chapter.

\chapter{Layer Partitioning}

\begin{figure}
\centering
\includegraphics[scale=.4]{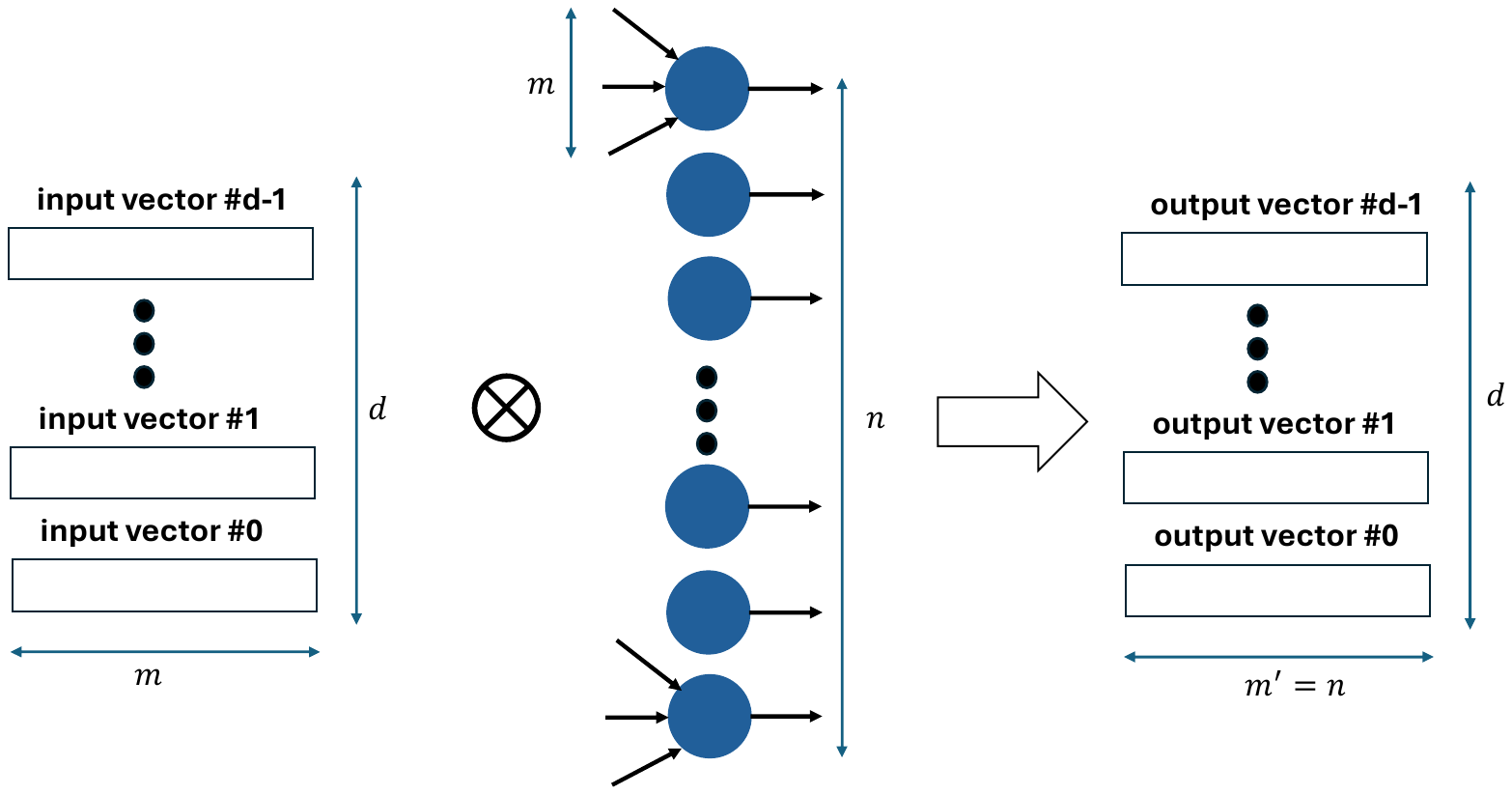}
\caption[Fully connected layer parallelism.]{Fully connected layer parallelism.}
\label{fig:connected_parallel}
\end{figure}

We will continue exploring how to partition ML models within individual layers. However, each layer type may have different partitioning dimensions. This means multiple parallelism techniques can be used, depending on the specific layer type.

\section{Fully Connected Layers}

A fully connected (FC) layer and its corresponding weights in a neural network can be represented as matrix-vector multiplications. In Figure \ref{fig:connected_parallel}, we note the \emph{d}, \emph{m}, and \emph{n} dimensions. The \emph{d} dimension represents the input samples to the network, where each sample is a vector with dimension \emph{m}. This input data vector is then multiplied with a weight matrix of dimensions $m \times n$, representing $n$ neurons with $m$ weighted inputs, producing a sequence of \emph{d} output vectors of dimension \emph{m'}, which is equivalent to \emph{n}. In addition to exploiting data parallelism by partitioning along the $d$ dimension as described in Chapter 2, FC layers provide additional intra-layer parallelizing opportunities by partitioning along the $m$ or $n$ dimensions.

Table \ref{tab:connectedSchemes} summarizes the different approaches for partitioning FC layers. Again, we categorize approaches based on inference and/or training, cloud and/or edge deployment, partitioning dimensions, computational benefits, communication requirements, memory benefits, and whether the approach protects private data.

The works in \cite{hadidi2020toward, stahl2019fully, mao2017modnn} focus on partitioning the inputs and outputs of FC layers through layer output and input partitioning. In layer output partitioning (LOP), the input vector of size \emph{m} is multiplied by a subset of neurons, evenly distributed among a chosen subset size of \emph{n} devices. After the summation of each subset is computed, an activation function is applied to each subset. Finally, each activated output is sent to the same device and concatenated to create the full output vectors of length \emph{n}. As a result of LOP, each device has an even memory footprint and communication cost related to the size of each subset \emph{n}. In layer input partitioning (LIP), \emph{m} is split into subsets and placed on separate devices. Then, each subset is multiplied by \emph{n} weights and sent to a final device that calculates the summation of all of the subsets and applies the activation function to each output vector of length \emph{n}. LOP can outperform LIP because LOP communicates less data overall as more values are set to zero when the activation function is applied before communication. Additionally, both methods decrease the total memory required on each device and increase the system's throughput as the inference tasks can be pipelined. However, this method does not protect data privacy, as all data is transmitted to the device that
holds the network’s input layer. 

LIP and LOP support the fusing of operations, where the partitioned outputs of an FC layer on which LOP has been applied can be directly fed as partitioned inputs to a subsequent LIP setup without having to assemble complete intermediate vectors, i.e., without the need to communicate between devices. Note that in the case of FC layers, such a fused LOP-LIP combination can, at maximum, encompass two layers.

In addition to input and output partitioning for inference, works such as \cite{osti_10404274} describe an approach for FC layer training that uses ideas from both federated learning and distributed training by partitioning fully connected layers across the \textit{m} and \textit{n} dimensions with the same number of layers as the original model on edge devices. The sub-models are then trained, and weight updates are shared, similar to FL, which in turn decreases the synchronization overhead as synchronization only needs to happen once after the sub-models are trained. This increases throughput in the system and allows networks to be fully trained on edge devices with limited memory. This approach protects private data as all input data is kept on the collected device.

While fully connected layers play an essential role in ML models, computer vision tasks have gained popularity, making convolutional layers ubiquitous. This leads to more challenges when determining how to run these computer vision networks on memory-constrained devices.

\begin{table*}[]
\caption[Partitioning schemes for fully connected layers.]{Partitioning schemes for fully connected layers.}
\makebox[\linewidth]{
\begin{tabulary}{.8\linewidth}{ c | c | c | c | c | c | c | c }  \makecell{} &
\makecell{Inf./\\train.} & \makecell{Cloud/\\Edge} & \makecell{part \\dim.} & 
\makecell{comp. \\benefit} & \makecell{comm. \\req.} & \makecell{memory \\benefit} & \makecell{privacy}\\
\hline
\hline
\makecell{Output-Based \\ Partitioning \\ \cite{hadidi2020toward, stahl2019fully, mao2017modnn}} & Inf. & Edge & $j$  & \makecell{throughput} & output &\makecell{output/\\weights} & -\\
\hline
\makecell{Input-Based\\ Partitioning \\ \cite{hadidi2020toward, stahl2019fully}} & Inf. & Edge & $i$ & throughput & input &\makecell{input/\\weights} & - \\
\hline 
\makecell{Hybrid\\ \cite{osti_10404274}} & Train. & Edge & $i$/$j$ & throughput & weights & \makecell{input/\\output/\\weights}  & x\\
\end{tabulary}
}
\label{tab:connectedSchemes}
\end{table*}

\section{Convolutional Layers}

\begin{figure}[!h]
\centering
\includegraphics[width=\textwidth]{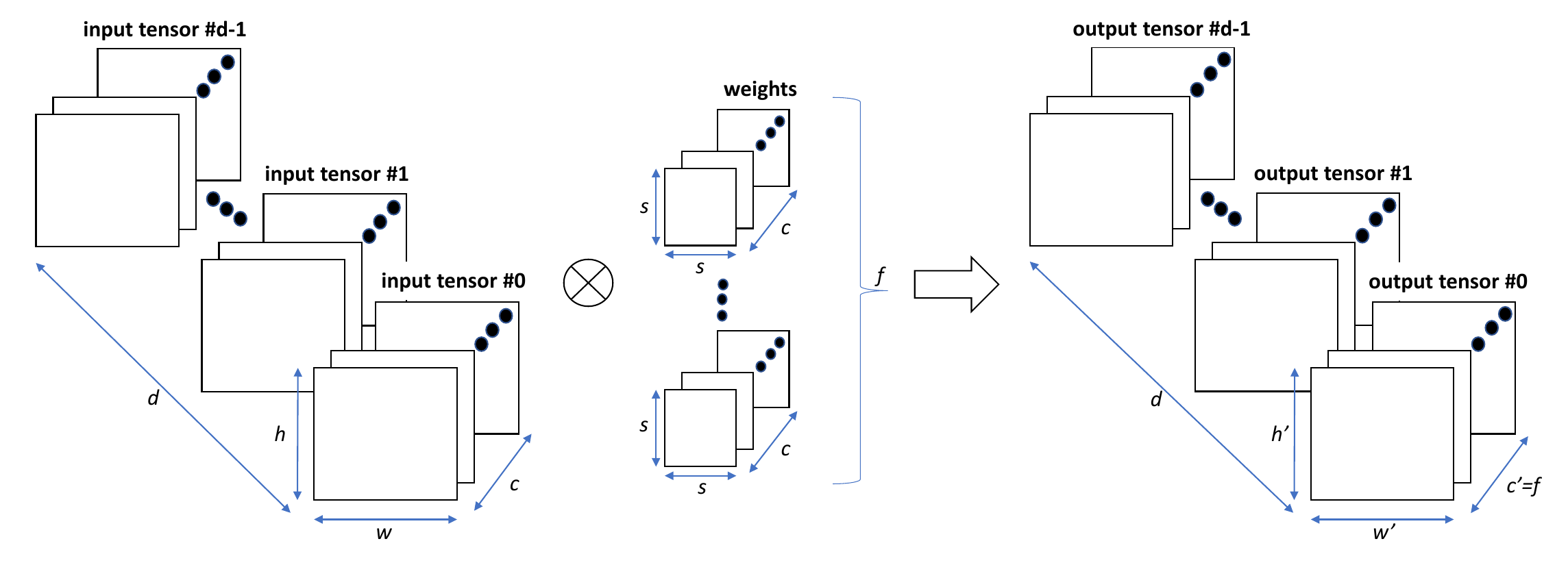}
\caption[Convolutional layer parallelism.]{Convolutional layer parallelism.}
\label{fig:conv_parallel}
\end{figure}

Figure \ref{fig:conv_parallel} describes the architecture of convolutional layers. Multiple filters \emph{f}, denoted with height and width \emph{s} and depth \emph{c}, are convolved with the input tensors to that layer, of width $w$, height \emph{h}, and channel \emph{c} dimensions, forming a so-called feature map of size $w \times h$ and depth \emph{c}. This produces an output tensor (feature map) of width \emph{w'}, height \emph{h'}, and depth $c'$. The size of the output feature map depends on the input width and height ($w \times h$), padding, and striding of how the filters are applied, while each filter produces one output channel, i.e., the number of output channels \emph{c'} is equal to the number of filters ($f$).

Table \ref{tab:convSchemes} summarizes partitioning strategies for convolutional layers. In general, we can distinguish strategies based on their focus on the partitioning of feature maps or filter weights.

\begin{table*}[]
\begin{center}
\caption[Partitioning schemes for convolutional neural networks.]{Partitioning schemes for convolutional neural networks.}
\begin{tabular}{ c | c | c | c | c | c | c | c} \makecell{} &
\makecell{Inf.\\/Train.} & \makecell{Cloud\\/Edge} & \makecell{part.\\ dim.} & 
\makecell{comp.\\benefit} & \makecell{comm. \\req.} & \makecell{memory \\ benefit} & \makecell{privacy}\\
\hline
\hline
\makecell{Feature Partitioned\\Inference \\\cite{mao2017modnn, zhao2018deepthings, zeng_coedge_2020, gao_edgesp_nodate, wang_edgeduet_2021}} & Inf. & Edge & $w$/$h$ & \makecell{latency} & \makecell{input/\\output} & \makecell{input/\\output} & x\\ 
\hline
\makecell{Weight Partitioned\\ Inference \\\cite{deyPartitioningCNNModels2018, stahl_deeperthings_2021}} & Inf. & Edge & $c$/$f$ & \makecell{latency} & \makecell{weights} & weights & x\\
\hline
\makecell{Weight Partitioned \\ Training \\\cite{chollet2017xception,drydenChannelFilterParallelism2019}} & Train. & Cloud & $c$/$f$ & \makecell{latency} & \makecell{weights} & weights & -\\
\end{tabular}
\label{tab:convSchemes}
\end{center}
\end{table*}

\subsection{Feature Map Partitioning}

A common strategy for partitioning convolutional layers is to tile the layer across the \emph{h} and \emph{w} dimensions. This exploits the inherent locality in convolutions, where each device processes one input tile to produce a corresponding output tile. MoDNN \cite{mao2017modnn} partitions convolutional layers in the \textit{h} or \textit{w} dimensions to minimize the need for nodes in the cluster to communicate. While this reduces data dependencies and the memory required to store intermediate feature map data, it maintains layer-by-layer execution, potentially causing network bottlenecks and lacking dynamic adaptation to varying computing demands. 

\begin{figure}[h!]
\centering
\includegraphics[scale = .5]{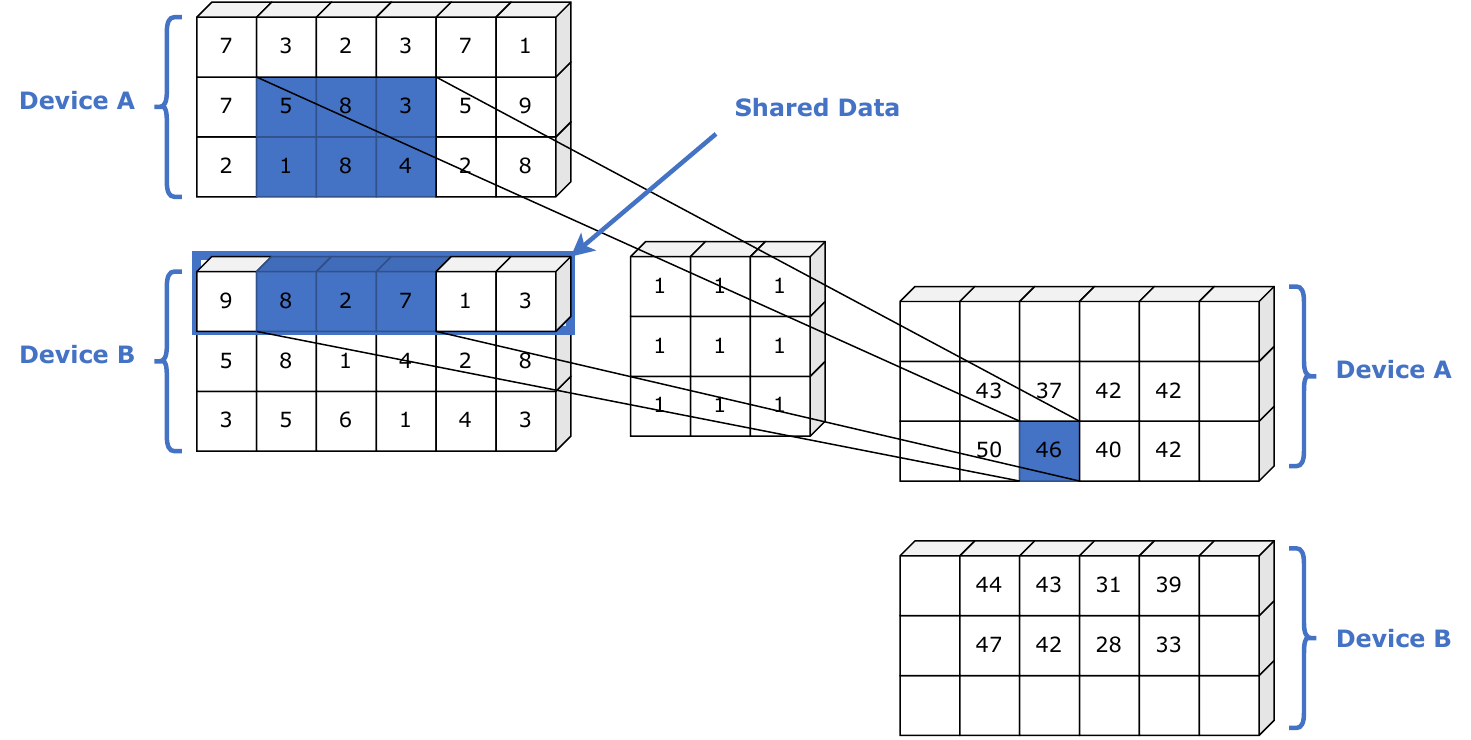}
\caption[Shared data in distributed convolutional operations.]{Shared data in distributed convolutional operations.}
\label{fig:shared_conv}
\end{figure}

Layer fusion, introduced in \cite{layerFusion}, aims to further reduce data transmission in a network. In layer fusion, the outputs of one layer of the network are sent directly as the inputs to the next layer of the network on the same device, bypassing the need to communicate intermediate feature map data between devices. However, since data regions overlap in convolutional operations, as shown in Figure \ref{fig:shared_conv}, the overlapping segments of the nodes must still be communicated. Consequently, device dependency increases, resulting in the reliance on communication from other devices for shared data. However, in layer fusion, data privacy is protected as most information is kept local on each edge device and not offloaded to other devices or the cloud for further processing.

DeepThings \cite{zhao2018deepthings} implements a fusing approach along with tile-based partitioning by dividing convolutional layers into independent tasks based on local regions for parallel execution. This approach reduces memory usage and communication overhead by fusing intermediate feature maps within edge nodes, leading to more efficient data transmission than MoDNN. In addition to the previously discussed works, a large amount of research has been done on the topic, such as introducing heterogeneous devices into the edge cluster \cite{zeng_coedge_2020, wang_edgeduet_2021, gao_edgesp_nodate} to maximize resource utilization and decrease the total latency of the system.

\subsection{Channel, Filter, and Weight Partitioning}
In convolutional neural networks (CNNs), the channel dimension (\textit{c}) is critical for achieving high accuracy especially in later layers of deep CNNs.\cite{simonyanVeryDeepConvolutional2015}. Therefore, network computational complexity increases as the \emph{c} dimension increases. As a result, partitioning convolutional layers in the channel dimension \textit{c} is a popular method that decreases the latency of a model and increases throughput. Partitioning in the \emph{c} dimensions splits both feature maps and filters to process a subset of channels on each device. This reduces memory requirements for both feature map and weight data, but requires communication and summation of the outputs produced by each partition/device to obtain output tensors. 

Alternatively, filter partitioning, or partitioning in the \emph{f} dimension simply assigns one complete filter to each partition/device, where each partition/device produces one channel of the output feature map, where channels only need to be assembled to form the complete map. This partitioning strategy allows for effective distribution of computation in weight- or filter-dominated layers across devices while minimizing communication and memory overhead, optimizing the execution of CNNs on resource-constrained systems.

Partitioning a network in both the \emph{c} and \emph{f} dimensions aims to further reduce the overhead incurred by partitioning in the \emph{h} and \emph{w} dimensions alone, reducing memory and communication overhead while enhancing efficiency \cite{deyPartitioningCNNModels2018, stahl_deeperthings_2021}. This approach considers resource constraints in edge clusters and demonstrates performance improvements in well-known CNNs. However, note that in contrast to feature map partitioning, similar to FC layer input-output partitioning, this work only allows for pairwise fusing of filter partitioned with channel partitioned layers, i.e. weight partitioning does not directly support arbitrary fusing of layers, therefore, compared to feature partitioning techniques, this partitioning scheme has a larger communication overhead.

The work in \cite{chollet2017xception} proposes Xception, a CNN architecture based on depthwise separable convolutions, which inherently allow for partitioning of depthwise and pointwise convolution operations across the channel (\emph{c}) and filter (\emph{f}) dimensions, respectively. This approach enhances parameter efficiency and model performance in image classification tasks while reducing computational complexity, facilitating faster training. By employing depthwise separable convolutions, Xception achieves a smaller model size due to fewer parameters, demonstrating effective channel partitioning strategies to optimize CNN efficiency. However, Xception does not decrease the communication overhead between devices in a cluster. In contrast, the work in \cite{drydenChannelFilterParallelism2019} emphasizes channel and filter parallelism to accelerate large-scale CNN training. This approach enables strong scaling, reduces communication overhead, and improves memory efficiency by distributing computation across channels and filters. While it introduces additional communication during training, volume and memory usage is optimized, further highlighting the importance of channel partitioning for enhancing CNN scalability and efficiency in training scenarios.

\section{Recurrent Layers}

\begin{figure} [!htb]
\centering
\includegraphics[scale = .5]{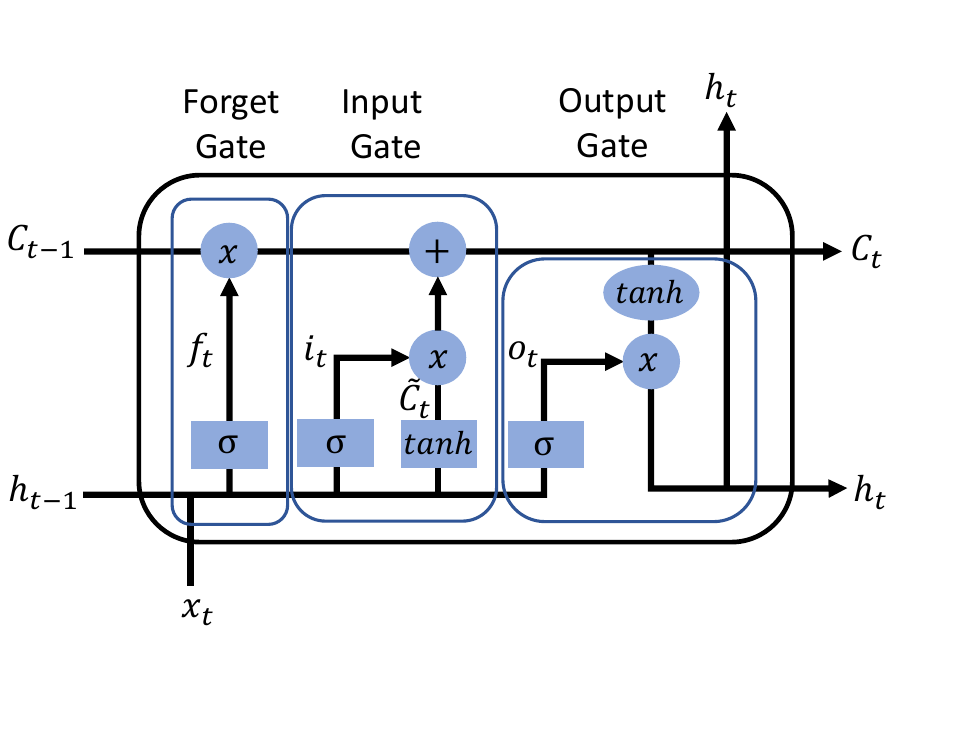}
\caption[Long Short-Term Memory (LSTM) cell.]{Long Short-Term Memory (LSTM) cell.}
\label{fig:lstm_cell}
\end{figure}

Recurrent Neural Networks (RNNs) represent a class of artificial neural networks designed explicitly for processing sequential data. Their architecture incorporates recurrent connections, enabling information to propagate across time steps. This recurrent structure can be conceptualized as an unrolled network, where each iteration receives input from the current element in the sequence and the hidden state of the previous iteration. This hidden state is a memory mechanism encoding information from past inputs and influencing the network's response to subsequent elements. However, RNNs are susceptible to the vanishing gradient problem, where gradients diminish as they backpropagate through time, hindering the network's ability to learn long-range dependencies within the sequence.

To address the limitations of RNNs, precisely the vanishing gradient problem, Long Short-Term Memory Networks (LSTMs) were introduced as shown in Figure \ref{fig:lstm_cell}. LSTMs, a specialized variant of RNNs, augment the architecture with a gated cell state mechanism. This cell state acts as a "long-term memory," allowing the network to retain information over extended periods. Three gates regulate the flow of information into and out of the cell state: the forget gate, which selectively discards irrelevant information; the input gate, which determines what new information to store; and the output gate, which controls the information retrieved from the cell state for generating the current output. The necessary equations used in an LSTM network are shown below, where the $W$ variables form matrices that represent the network's weights.

\begin{equation}
    i_t = \sigma(x_tU^i + h_{t-1}W^i)
\end{equation}
\begin{equation}
    f_t = \sigma(x_tU^f + h_{t-1}W^f)
\end{equation}
\begin{equation}
    o_t = \sigma(x_tU^o + h_{t-1}W^o) 
\end{equation}
\begin{equation}
    \Tilde{C}_t = tanh(x_tU^g + h_{t-1}W^g)
\end{equation}
\begin{equation}
    C_t = \sigma(f_t * C_{t-1} + i_t * \Tilde{C}_t)
\end{equation}
\begin{equation}
    h_t = tanh(C_t) * o_t
\end{equation}

By mitigating the vanishing gradient problem, LSTMs excel at capturing long-range dependencies in sequential data, making them superior to traditional RNNs for tasks requiring extended memory, such as natural language processing and time series prediction \cite{hochreiter1997long}.

 To optimize performance and scalability, partitioning LSTM layers in ML involves splitting the LSTM computations across different dimensions, such as the forget, input, and output gates. Table \ref{tab:lstm_layers} details partitioning methods of recurrent layers.

\begin{table*}[]
\begin{center}
\caption[Existing works on RNN partitioning.]{Existing works on RNN partitioning.}
\begin{tabular}{ c | c | c | c | c}
 & \makecell{Inf./\\Train.} & \makecell{Cloud/\\Edge} & \makecell{part.\\dim.} & privacy \\
\hline
\hline
\makecell{Gate \\ Partitioned\\ \cite{chang2015recurrent,nurvitadhi2016accelerating}} & \makecell{Inf./\\Train.} & Edge & $\sigma$/$\tanh$ & x \\
\hline
\makecell{Weight \\Partitioned\\ \cite{kuchaiev2017factorization, cao2019efficient, kwon2020scalable}} & \makecell{Inf./Train.} & Edge & $W$ & x \\
\hline
\makecell{Model \\Partitioned\\ \cite{kuchaiev2017factorization}} & \makecell{Train.} & Cloud & \makecell{$W$/$x$/$h$} &  \\
\end{tabular}
\end{center}
\label{tab:lstm_layers}
\end{table*}

\subsection{Gate-Based Partitioning}

One effective way to partition LSTM layers is through gate-wise decomposition. Each LSTM gate (forget, input, and output) can be treated as an independent computational unit. This approach allows for parallelization and distribution of computations. For instance, the LSTM layer can be divided into sub-layers corresponding to each gate. In this setup, the forget gate sub-layer handles computations related to forgetting information from the previous cell state, and the input gate sub-layer manages input modulation and decides what new information to store. The output gate sub-layer controls the output generation. By partitioning in this manner, each sub-layer can operate independently and efficiently utilizing hardware resources in distributed systems. Additionally, parameter sharing across these sub-layers can be leveraged to reduce memory footprint and increase training speed. This partitioning strategy optimally distributes the workload, enhancing the scalability and performance of LSTM networks in ML tasks. 

Gate partitioning has been applied to custom hardware implementation of FSMs on FPGAs \cite{chang2015recurrent}. Due to the recurrent nature of LSTMs, traditional hardware does not allow for maximum performance. CPUs do not offer large amounts of parallelism, and small RNNs do not fully benefit from the parallelization of GPUs. Therefore, a case is made for specialized hardware to run inference tasks on these models. To increase parallelization in computation tasks, two sigmoid and one tanh gates are implemented to allow equations used in the LSTM network to occur in parallel if no dependencies exist. The work in \cite{nurvitadhi2016accelerating} verifies that the FPGA approach is more efficient than CPU and GPU-based approaches due to the ability to extract fine-grained parallelism in LSTM modules.

\subsection{Weight-Based Partitioning}

The weight-based partitioning methods during inference focus on achieving scalable RNN acceleration on FPGA platforms. The work in \cite{kwon2020scalable} introduces three levels of parallelism—matrix-level, operation-level, and layer-level—to optimize RNN processing across multiple FPGAs. Matrix-level parallelism (weight-based partitioning) simply takes matrix-vector multiplications, where the matrix is the weight matrix, $W$, and partitions these multiplications into independent execution units based on either the rows or columns of $W$. Operation-level parallelism takes entire matrix-vector multiplications in a layer and executes them on different functional units. For example, the multiplications involving $x_t$ and $h_t$ would be placed on different functional units and executed in parallel. Layer-level parallelism is placing each layer in the network on separate FPGAs. These parallelism approaches also include analyzing dependencies within RNNs and implementing software pipelining to enhance hardware utilization. Additionally, the work in \cite{cao2019efficient} introduces Bank-Balanced Sparsity (BBS) to achieve high accuracy and hardware efficiency by partitioning weight matrix rows into equally sized banks and applying fine-grained pruning. Moreover, the work in \cite{kuchaiev2017factorization} tackles the challenge of training large LSTM networks by proposing Factorized LSTM (F-LSTM), which decomposes the LSTM weight matrix ($W$) into the product of two smaller matrices ($W_1$ and $W_2$), reducing the total parameter count and computational complexity. Together, these works showcase diverse weight partitioning strategies tailored for RNN acceleration on FPGA platforms, emphasizing parallelism exploitation, hardware awareness, and software-hardware co-design to optimize performance based on specific model requirements and hardware constraints.

\subsection{Model Partitioning}

Layer-based partitioning in \cite{kwon2020scalable} described in Section 3.3.2 realizes a form of model parallelism. In addition, the work in \cite{kuchaiev2017factorization} also proposes an alternative model partitioning scheme for training LSTMs called Group LSTM (G-LSTM). This scheme partitions the LSTM cell, inputs ($x_t$), and hidden states ($h_t$) into independent groups, each operating on a subset of features with its own set of parameters. G-LSTM enables parallel processing during training and reduces the overall parameter count. The G-LSTM model can be interpreted as an ensemble of smaller LSTM models operating on different feature subsets concatenated to preserve feature independence. Both G-LSTM and F-LSTM (discussed in Section 3.3.2) in \cite{kuchaiev2017factorization} significantly reduce parameter counts and training times while maintaining accuracy, facilitating the training of more extensive LSTM networks on constrained hardware resources for improved model complexity and performance exploration.

\chapter{Challenges and Future Directions}
Distributed learning holds immense promise for scaling and optimizing machine learning applications; however, several critical challenges must be addressed to realize its full potential and ensure its effective deployment in practical settings.
One primary challenge is communication overhead, particularly in resource-constrained edge and mobile environments. Frequently, data exchange between distributed devices during training or inference can lead to bottlenecks and latency issues. Optimizing communication protocols, implementing efficient data compression techniques, and developing effective model partitioning strategies are crucial to mitigate this challenge.

Another significant hurdle is system heterogeneity within distributed learning environments. These systems often involve diverse devices with varying computational capabilities, memory capacities, and communication bandwidths. Addressing this heterogeneity requires the development of adaptive and robust algorithms capable of efficiently distributing workloads and gracefully handling device failures or communication disruptions.
Moreover, ensuring data privacy and security remains paramount, particularly in scenarios like fully distributed ML, where data privacy is a top concern as potentially private data needs to be communicated between devices.

Efficient resource management and scheduling across distributed devices present additional complexities. This involves optimizing the allocation of computational resources, considering factors like device availability, energy consumption, communication costs, and task deadlines. Developing intelligent resource management and scheduling algorithms is crucial for optimizing system performance and reliability.

As ML models become increasingly complex, model optimization and compression emerge as essential strategies to ensure efficient distributed execution. Techniques such as model pruning, quantization, and knowledge distillation enable the reduction of model size and computational requirements while compromising accuracy. Combining these methods with the current distributed ML techniques described in this report would allow for faster inference times due to increased throughput and lower latency.

Looking ahead, several promising future directions can further advance distributed learning. Developing adaptive and dynamic partitioning strategies that respond to changing system conditions, leveraging specialized hardware accelerators for specific tasks or layer types, and fostering standardization and interoperability through established protocols and frameworks are critical areas for future exploration and innovation.

By systematically addressing these challenges and actively pursuing these promising avenues, distributed learning will continue to evolve and play a transformative role in shaping the future landscape of machine learning applications.

\chapter{Summary and Conclusions}
This survey has explored the multifaceted landscape of distributed learning, highlighting its potential to address the growing computational demands of modern machine learning models. We have examined the core principles of data and model parallelism, analyzing how models are strategically partitioned across diverse devices and environments to optimize resource utilization and performance. We have shed light on the critical trade-offs between computational efficiency, communication overhead, and memory constraints by delving into various partitioning schemes for different layer types, including fully connected, convolutional, and recurrent layers.

Through the lens of distributed learning, we have gained valuable insights into optimizing model training and inference processes while safeguarding data security. The challenges and future directions discussed underscore the need for continued research and development in communication optimization, handling system heterogeneity, and protecting private data.

Traditionally, distributed learning strategies have been applied in different contexts, such as on the edge or within the cloud. However, proposed solutions to partitioning these networks share the same design space and contain many similarities. Therefore, distributed learning can unlock new possibilities for efficient, scalable, and privacy-aware machine learning applications across various domains by fostering collaboration and innovation within this field.

\section*{Acknowledgments}		
We would like to thank Kamyar Mirzazad Barijough for providing some of the initial figures and ideas for this report.

%



\addcontentsline{toc}{chapter}{Bibliography}
\bibliographystyle{IEEEtranN}  
\bibliography{references.bib}

\end{document}